\documentclass{article}
\usepackage{spconf,amsmath,graphicx,cite}
\usepackage{mathtools}
\usepackage{tikz}
\usetikzlibrary{er,positioning,fit,trees,shapes,decorations}
\usepackage[font=small,skip=0pt]{caption}

\title{Neural-FST Class Language Model for End-to-End Speech Recognition}

\name{
\parbox{\linewidth}{\centering
    Antoine Bruguier$^*$,
    Duc Le$^*$,
    Rohit Prabhavalkar$^\dag$,
    Dangna Li,
    Zhe Liu,
    Bo Wang,
    Eun Chang,
    Fuchun~Peng,
    Ozlem Kalinli,
    Michael L. Seltzer\thanks{$^*$Equal contribution. $^\dag$Work performed while at Facebook. We'd like to thank Jun Liu and Frank Seide for their helpful comments and suggestions.}}}
\address{Facebook AI, USA\\ \small{\texttt{tonybruguier@fb.com}}
}
\begin{document}
\ninept
\maketitle
\begin{abstract}
We propose Neural-FST Class Language Model (NFCLM) for end-to-end speech recognition, a novel method that combines neural network language models (NNLMs) and finite state transducers (FSTs) in a mathematically consistent framework. Our method utilizes a background NNLM which models generic background text together with a collection of domain-specific entities modeled as individual FSTs. Each output token is generated by a mixture of these components; the mixture weights are estimated with a separately trained neural decider. We show that NFCLM significantly outperforms NNLM by 15.8\% relative in terms of Word Error Rate. NFCLM achieves similar performance as traditional NNLM and FST shallow fusion while being less prone to overbiasing and 12 times more compact, making it more suitable for on-device usage.
\end{abstract}
%



\section{Introduction}
\label{sec:introduction}
End-to-end (E2E) automatic speech recognition (ASR) models are becoming increasingly popular, especially for on-device applications, due to their compact size and competitive transcription accuracy~\cite{HeSainathPrabhavalkarEtAl19, KimLeeGowdaEtAl19}.
Unlike conventional hybrid ASR systems~\cite{MorganBourlard95}, which consist of separately trained acoustic (AM) and language models (LM), E2E models are trained jointly on paired acoustic-text data.
If paired data are available for all domains of interest, then E2E models can be made more robust by training on diverse data~\cite{NarayananPrabhavalkarChiuEtAl19}.
However, in many practical situations it is unreasonable to expect that paired data are available for all domains, either because such data collection would be prohibitively expensive or impractical.
In order to ensure that E2E ASR systems can be suitable replacements for conventional hybrid systems, we are presented with two challenges: rapidly adapting the E2E ASR system to support a new domain of interest, for which only text data might be available; and, ensuring that E2E ASR models can correctly recognize rare words (typically named entities), which might be unseen during training.
The first of these challenges has been addressed in previous work through \emph{language model fusion}~\cite{GulcehreFirhatXuEtAl15, KannanWuNguyenEtAl18, SriramJunSatheeshEtAl17}.
The second challenge, recognizing rare words, has only been most successfully addressed in the setting where we have a set of entities (e.g., user's contacts, song playlists, etc...) through \emph{contextual biasing}~\cite{ZhaoSainathRybachEtAl19,Le2021deepshallow,le21_interspeech}.

For this paper, we focus on the problem of developing an E2E ASR model which can perform well on multiple domains, where a subset of domains are only available as a set of context-free grammar (CFG) rules~\cite{HopcroftMotwaniUllman01}. 
This setting presents an important challenge when building speech-enabled voice assistants that must be able to support downstream spoken language understanding (SLU) systems which constantly expand coverage by serving new domains.
One approach to this problem is to generate sentences from the underlying domain grammar, and then generate the corresponding acoustic sequences using a text-to-speech (TTS) system (e.g., ~\cite{HeSainathPrabhavalkarEtAl19, KaritaWatanabeIwataEtAl19, RosenbergZhangRamabhadranEtAl19}).
However, this approach is computationally expensive, both the TTS synthesis process as well as the effort required to re-train the E2E model, which makes it harder to rapidly adapt to a new domain.
In this work we focus on using a fixed pre-trained E2E ASR model and improving the LM used during decoding.

The simplest approach to building neural LMs which cover multiple domains, some of which are represented as CFGs, is to generate domain-specific data by substituting non-terminal symbols~\cite{GalescuRinggerAllen98, WangMahajanHuang00}.
This data can then be pooled together with the ``background" text to train a single neural LM with wider coverage.
While this solution (which serves as one of our baselines) is simple to implement, it requires model re-training to ingest new patterns or entities. Extending this solution to work without re-training is not trivial for \emph{neural LMs}; some early work along this direction was explored in \cite{le21_interspeech}.
An alternative approach, employed extensively in hybrid ASR systems~\cite{MorganBourlard95} with \emph{n-gram LMs}, utilizes class-based non-terminals in the decoder graph to represent classes of interest (e.g., contact names, songs, artists, etc...); these can then be replaced on-the-fly using separate finite state transducers (FSTs) \cite{MohriPereiraRiley02} which can be compiled and personalized for each user~\cite{SchalkwykHetheringtonStory03, DixonHoriKashioka12, AleksicAllauzenElsonEtAl15}.
The corresponding solution for E2E models typically involves shallow fusion contextual biasing~\cite{ZhaoSainathRybachEtAl19, HaynorAleksic20,Le2021deepshallow, le21_interspeech}.
The FST-based biasing technique has two disadvantages: it is somewhat challenging to set the weights in the biasing FST, which are typically hand-tuned on a development set; and, there is a danger of \emph{overbiasing} phrases if the biasing weights are set to inappropriate values, although this can be mitigated somewhat by only allowing the biasing to be applied in specific contexts~\cite{HeSainathPrabhavalkarEtAl19}.
One possibility is to use the ASR model's output to decide when to apply biasing~ \cite{appleclasslms} but this means that the LM is tied to the ASR model.

In this work, we propose Neural-FST Class Language Model (NFCLM), an approach that builds a factorized LM consisting of a \emph{background} neural LM~\cite{BengioDucharmeVincentEtAl03, MikolovKombrinkBurgetEtAl11}, which is intended to have broad coverage for general text data, along with several separate domain-specific LMs.
Each word in the sentence is generated by this mixture of LM components weighted by a separately estimated neural \emph{decider}.
While our work is similar to previous work on mixture-of-expert LMs~\cite{ShazeerMirhoseiniMaziarzEtAl17, IrieKumarNirschlEtAl18}, a crucial aspect of our proposed model is that the domain specific LMs are represented using FSTs, analogous to their use in shallow fusion biasing~\cite{ZhaoSainathRybachEtAl19,Le2021deepshallow,le21_interspeech}.
However, in our work, these are integrated in a purely probabilistic model without the need for hand-tuning of biasing weights.
Leveraging FSTs to represent the domain-specific entities makes it easier to ingest new entities into the LM, which allows for rapid adaptation to a new domain or rare named entities.
The proposed NFCLM is used in shallow fusion with internal LM subtraction~\cite{VarianiRybachAllauzenEtAl20} during decoding. Our experiments demonstrate that NFCLM outperforms NNLM in modeling rare named entities, reducing Word Error Rate (WER) by 15.8\% relative on an entity-heavy evaluation set. Compared to the traditional NNLM and FST shallow fusion, NFCLM achieves similar WER while being less prone to overbiasing and 12 times more compact, making it a more suitable choice for on-device ASR.

\begin{figure*}
\resizebox{\textwidth}{!}{%
\begin{tikzpicture}[
scale=0.5,
graynode/.style={rectangle, draw=black!100, fill=black!5, thin, minimum size=5mm, align=left},
rednode/.style={rectangle, draw=red!100, fill=red!5, thin, minimum size=5mm, align=left},
bluenode/.style={rectangle, draw=blue!100, fill=blue!5, thin, minimum size=5mm, align=left},
greennode/.style={rectangle, draw=green!100, fill=green!5, thin, minimum size=5mm, align=left},
prunednode/.style={rectangle, draw=black!15, fill=black!0, thin, minimum size=5mm, align=left},
]
\node[] (start0)                  {
};
\node[] (start1)     [below=10pt of start0]             {};

\node[] (play0)  [right=30pt of start0]                  {
$\mathbf{h^w}=$\_play};
\node[bluenode] (play1)  [below=10pt of play0] {
$\mathbf{h^\text{decider}}=$\_play};

\node[draw, draw=black!50, fit=(play0) (play1)](playrosie) {
    };

\node[] (playro0) [right=50pt of play0]                 {
\_play,\_ro};
\node[bluenode] (playro1)  [below=10pt of playro0] {
\_play,\_ro};
\node[rednode] (playro2)  [below=10pt of playro1] {
\_play,@song \\
$s(@song)=1$};
\node[greennode] (playro3)  [below=10pt of playro2] {
\_play,@artist \\
$s(@artist)=2$};

\node[draw, draw=black!50, fit=(playro0) (playro1) (playro2) (playro3)](playro) {};

\node[] (playrosie0) [right=50pt of playro0]                 {
\_play,\_ro,sie};

\node[bluenode] (playrosie1)  [below=10pt of playrosie0] {
\_play,\_ro,sie};

\node[rednode] (playrosie2)  [below=10pt of playrosie1] {
\_play,@song \\
$s(@song)=2$};

\node[draw, draw=black!50, fit=(playrosie0) (playrosie1) (playrosie2)](playrosie) {
    };

\node[] (playrosieby0) [right=50pt of playrosie0]                 {
\_play,\_ro,sie\_by};

\node[bluenode] (playrosieby1)  [below=10pt of playrosieby0] {
\_play,\_ro,sie,\_by};

\node[bluenode] (playrosieby2)  [below=40pt of playrosieby1] {
\_play,@song,\_by};

\node[draw, draw=black!50, fit=(playrosieby0) (playrosieby1) (playrosieby2)](playrosieby) {
    };

\node[] (playrosiebybrowne0) [right=50pt of playrosieby0]                 {
\_play,\_ro,sie,\_by,\_browne};

\node[bluenode] (playrosiebybrowne1)  [below=10pt of playrosiebybrowne0] {
\_play,\_ro,sie,\_by,\_browne};

\node[prunednode] (playrosiebybrowne2)  [below=10pt of playrosiebybrowne1] {
\textcolor{gray}{\_play,\_ro,sie,\_by,@artist} \\
\textcolor{gray}{$s(@artist)=1$}};

\node[prunednode] (playrosiebybrowne3)  [below=10pt of playrosiebybrowne2] {
\textcolor{gray}{\_play,@song,\_by,\_browne}};

\node[greennode] (playrosiebybrowne4)  [below=10pt of playrosiebybrowne3] {
\_play,@song,\_by,@artist \\
$s(@artist)=1$};

\node[draw, draw=black!50, fit=(playrosiebybrowne0) (playrosiebybrowne1) (playrosiebybrowne2) (playrosiebybrowne3) (playrosiebybrowne4)](playrosiebybrowne) {
    };

\draw[->] (start1.east) -- (play1.west) node [midway, fill=white] {{$P_1$}};
\draw[->] (play1.east) -- (playro1.west);
\draw[->] (play1.east) -- (playro2.west) node [midway, fill=white] {{$P_2$}};
\draw[->] (play1.east) -- (playro3.west);
\draw[->] (playro1.east) -- (playrosie1.west);
\draw[->] (playro2.east) -- (playrosie2.west) node [midway, fill=white] {{$P_3$}};
\draw[->] (playrosie1.east) -- (playrosieby1.west);
\draw[->] (playrosie2.east) -- (playrosieby2.west) node [midway, fill=white] {{$P_4$}};
\draw[->] (playrosieby1.east) -- (playrosiebybrowne1.west);
\draw[->] (playrosieby1.east) -- (playrosiebybrowne2.west);
\draw[->] (playrosieby2.east) -- (playrosiebybrowne3.west);
\draw[->] (playrosieby2.east) -- (playrosiebybrowne4.west) node [midway, fill=white] {{$P_5$}};

\end{tikzpicture}
}
\caption{NFCLM's expanded dynamic FST after consuming the symbol sequence $\mathbf{h^w} = \{\_play, \_ro, sie, \_by, \_browne\}$. The decider histories $\mathbf{h^\text{decider}}$ left inside the beam after each extension are displayed on the state label. Here we assume that there are two classes, $@song$ (red) and $@artist$ (green), see Fig.~\ref{fig:bias_fst}, in addition to the background states (blue). In the last extension, two paths are dropped from the beam (gray). As an illustration, $P_1=P_D(@bg|\{\})P_{@bg}(\_play|\{\})$,
$P_2=P_D(@song|\{\_play\})P_{@song}(\_ro|\{\})$,
$P_3=P_{@song}(sie|\{\_ro\})$,
$P_4=P_D(@bg|\{\_play,@song\})P_{@bg}(\_by|\{\_play,\_ro,sie\})$, and
$P_5=P_D(@artist|\{\_play,@song,\_by\})P_{@artist}(\_browne|\{\})$.
}
\label{fig:dfst}
\end{figure*}
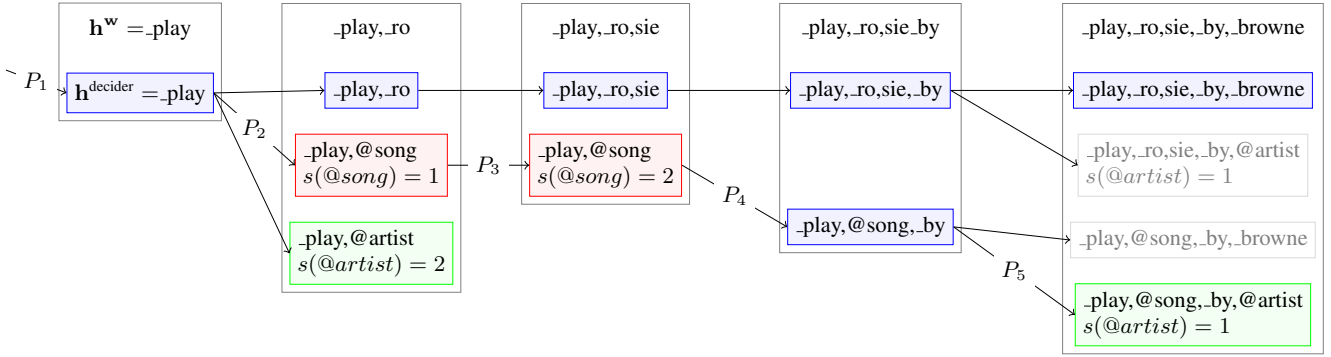

\section{Neural-FST Class Language Model (NFCLM)}
\label{sec:nfclm}

\subsection{Theoretical Formulation}
\label{sec:proposed_method_theory}
Let us assume we have an end-to-end ASR model trained to output symbols over some sub-word vocabulary, $\mathcal{V}$ (WordPieces~\cite{KudoRichardson18}, in this work). We introduce another set $\mathcal{C}$ consisting of non-terminal background (\texttt{@bg}) and non-background (e.g., \texttt{@song}, \texttt{@artist}) classes, which is disjoint from $\mathcal{V}$. For convenience, we define a special symbol $\epsilon$ to represent the continuation of an existing non-background class, and an expanded set $\mathcal{C}_\epsilon = \mathcal{C}~\cup~\{\epsilon\}$.

Given a length-n output symbol history $\mathbf{h^w} \in \mathcal{V}^n$, let $\mathbf{h^c} \in \mathcal{C}_\epsilon^n$ be a possible class alignment of $\mathbf{h^w}$, which indicates that the symbol $h^w_i$ is generated by the class $h^c_i$. Then, we can define the NFCLM probability for the next output symbol $w$ by marginalizing over all possible alignments and all possible classes $c$ which can generate $w$:


\begin{align}
    P(w | \mathbf{h^w}) &= \sum_{\mathbf{h^c} \in \mathcal{C}_\epsilon^n} P(\mathbf{h^c} | \mathbf{h^w}) \sum_{c \in \mathcal{C}_\epsilon} P(w | c, \mathbf{h^c}, \mathbf{h^w}) P(c | \mathbf{h^c}, \mathbf{h^w}) \label{eq:clm}
\end{align}

As can be seen in Eq.~\eqref{eq:clm}, the overall NFCLM consists of the \emph{class component probability} $P(w | c, \mathbf{h^c}, \mathbf{h^w})$, the \emph{class emission probability}
$P(c | \mathbf{h^c}, \mathbf{h^w})$, and the \emph{alignment probability} $P(\mathbf{h^c} | \mathbf{h^w})$. Note that unlike previous work on class-based LMs~\cite{BrownDellaPietraDesouzaEtAl92}, we do not assume that each output symbol belongs to a fixed class in $\mathcal{C}_\epsilon$; this is especially important since we model sub-word units which by their nature are shared across all classes.

\subsubsection{Class Component Probability: $P(w | c, \mathbf{h^c}, \mathbf{h^w})$}
\label{sec:component_lms}



The goal of this distribution is to estimate the probability of producing the output symbol $w$, given the output symbol history $\mathbf{h^w}$, its alignment $\mathbf{h^c}$, and the assumption that $w$ is generated by the class $c$. Let $f(\mathbf{h^c}, c)$ be a function that returns the last non-epsilon class in the sequence $\{\mathbf{h^c}, c\}$ if there is any (i.e., the class we are currently in), or $\epsilon$ otherwise (i.e., no class has been emitted yet). Then we can formally define the class component probability:

\begin{equation}
    P(w | c, \mathbf{h^c}, \mathbf{h^w}) =
\begin{cases}
    P_\text{@bg}(w | \mathbf{h^w}) & \text{if } f(\mathbf{h^c}, c) = @bg \\
    P_{f(\mathbf{h^c}, c)}(w | \mathbf{h^c}, \mathbf{h^w}) & \text{else if } f(\mathbf{h^c}, c) \neq \epsilon \\
    0 & \text{otherwise} \label{eq:class_prob}
\end{cases}
\end{equation}

As can be seen in Eq.~\eqref{eq:class_prob}, we model the background and non-background class LMs differently. The background class LM, $P_\text{@bg}(w | \mathbf{h^w})$, models generic open-domain text, thus it does not take $\mathbf{h^c}$ into account and can be viewed as a standard neural LM over $\mathcal{V}$. For non-background classes, our goal is to allow easy injection of new entities into the model without re-training, thus we represent them as FSTs built from class-specific data. Unlike the FST in \cite{ZhaoSainathRybachEtAl19} which used hand-tuned biasing weights, our FST is constructed so that the weights can be interpreted as probabilities, i.e., the weights of all outgoing arcs from any given state add up to 1.0. More formally, these FSTs estimate $P_c(w | \mathbf{h^c}, \mathbf{h^w}) = P_c(w | \mathbf{h_c^\text{prefix}})$, where $\mathbf{h_c^\text{prefix}}$ is defined as the last symbols in $\mathbf{h^w}$ that are generated by the non-background class $c$ or its continuation. For example, given $\mathbf{h^w} = \{\_play, \_ro, sie\}$ and $\mathbf{h^c} = \{@bg, @song, \epsilon\}$, then $\mathbf{h_\text{@song}^\text{prefix}} = \{\_ro, sie\}$ and $\mathbf{h_\text{@artist}^\text{prefix}} = \{\}$.

Let us also define $P_c(\phi|\mathbf{h_c^\text{prefix}})$ as the probability of exiting from the FST of the non-background class $c$ after consuming $\mathbf{h_c^\text{prefix}}$. This probability will be zero if we end up in a non-final FST state after consuming $\mathbf{h_c^\text{prefix}}$ and non-zero otherwise. For convenience, let $P_\text{@bg}(\phi|\mathbf{h^w}) = 1$ so that $P(\phi | c, \mathbf{h^c}, \mathbf{h^w})$ is defined for both the background and non-background classes.

\subsubsection{Class Emission Probability: $P(c | \mathbf{h^c}, \mathbf{h^w})$}
\label{sec:decider}

The goal of this distribution is to predict which class $c$ is likely to come next given a history of output symbols $\mathbf{h^w}$ and its alignment $\mathbf{h^c}$. This is computed primarily with a \emph{decider} neural network trained separately to estimate $P_D(c | \mathbf{h^c}, \mathbf{h^w}) = P_D(c | \mathbf{h^\text{decider}})$, where $c \in \mathcal{C}$ and the decider history $\mathbf{h^\text{decider}}$ is the same as $\mathbf{h^c}$, but with all $\epsilon$ removed and $@bg$ tokens replaced with the corresponding WordPiece. For example, if $\mathbf{h^w} = \{\_play, \_ro, sie\}$ and $\mathbf{h^c} = \{@bg, @song, \epsilon\}$, then $\mathbf{h^\text{decider}} = \{\_play, @song\}$. By formulating the decider this way, we are able to condition on both regular output symbols as well as the class labels. We can then formally define the class emission probability:

\begin{equation}
    P(c | \mathbf{h^c}, \mathbf{h^w}) =
\begin{cases}
    1 - P(\phi | c, \mathbf{h^c}, \mathbf{h^w}) & \text{if } c = \epsilon \\
    P(\phi | c, \mathbf{h^c}, \mathbf{h^w}) P_D(c | \mathbf{h^c}, \mathbf{h^w}) & \text{otherwise} \label{eq:class_emission}
\end{cases}
\end{equation}

From Eq.~\eqref{eq:class_emission} and the exit probability's definition in Section~\ref{sec:component_lms}, we see that the probability of emitting $\epsilon$ is always 0 unless we are in a non-final state of a non-background class FST.

\subsubsection{Alignment Probability: $P(\mathbf{h^c} | \mathbf{h^w})$}
\label{sec:alignment_prob}

The goal of this distribution is to estimate the probability that the output symbol sequence $\mathbf{h^w}$ corresponds to the alignment $\mathbf{h^c}$. Let us define $\mathbf{h^w_{:k}}$ and $\mathbf{h^c_{:k}}$ as the sub-sequences containing the first $k$ elements in $\mathbf{h^w}$ and $\mathbf{h^c}$, respectively. Then, the alignment probability can be expanded as follows using Bayes' rule:

\begin{align}
    P(\mathbf{h^c} | \mathbf{h^w}) &= P(h^c_1, h^c_2, \dots, h^c_n | h^w_1, h^w_2, \dots, h^w_n) \\
              &= P(h^c_1) P(h^c_2 | \mathbf{h^c_{:1}}, \mathbf{h^w_{:1}}) \dots P(h^c_n | \mathbf{h^c_{:n-1}}, \mathbf{h^w_{:n-1}}) \label{eq:alignment}
\end{align}

As shown in Eq.~\eqref{eq:alignment}, the alignment probability can be computed from the class emission probability of each label in the sequence, using Eq.~\eqref{eq:class_emission}. In practice, we make use of memoization to avoid re-computing this probability each time. Note that $P(h^c_1)$ corresponds to the class emission probability of the first symbol, when both $\mathbf{h^c}$ and $\mathbf{h^w}$ are empty: $P(c) = P(c | \mathbf{h^c}=\{\}, \mathbf{h^w}=\{\})$.

\subsection{Practical Implementation}
\label{sec:proposed_method_implementation}
\subsubsection{Beam Search Approximation}
\label{sec:decoding}

The number of possible alignment paths $\mathbf{h^c}$ grows exponentially with the length of the output symbol history $\mathbf{h^w}$, thus the computation of Eq.~\eqref{eq:clm} is intractable. As with many ASR decoding algorithms, we can approximate this equation with beam search:

\begin{equation}
    P(w | \mathbf{h^w}) \approx \sum_{\mathbf{h^c} \in \mathcal{B}_\mathbf{h^w}} P(\mathbf{h^c} | \mathbf{h^w}) \sum_{c \in \mathcal{C}_\epsilon} P(w | c, \mathbf{h^c}, \mathbf{h^w}) P(c | \mathbf{h^c}, \mathbf{h^w}) \label{eq:clm_beam}
\end{equation}

\noindent where the beam $\mathcal{B}_\mathbf{h^w}$ contains the most probable alignments of $\mathbf{h^w}$. In this work, we implement $\mathcal{B}_\mathbf{h^w}$ as a soft beam that retains up to $N=100$ alignments whose log-probability difference compared to the best alignment in the beam is at most $\delta=30$. Through this approximation, we can now compute $P(w | \mathbf{h^w})$ efficiently.




\subsubsection{Dynamic FST}
\label{sec:dynamic_fst}

\begin{figure}
\includegraphics[width=\columnwidth]{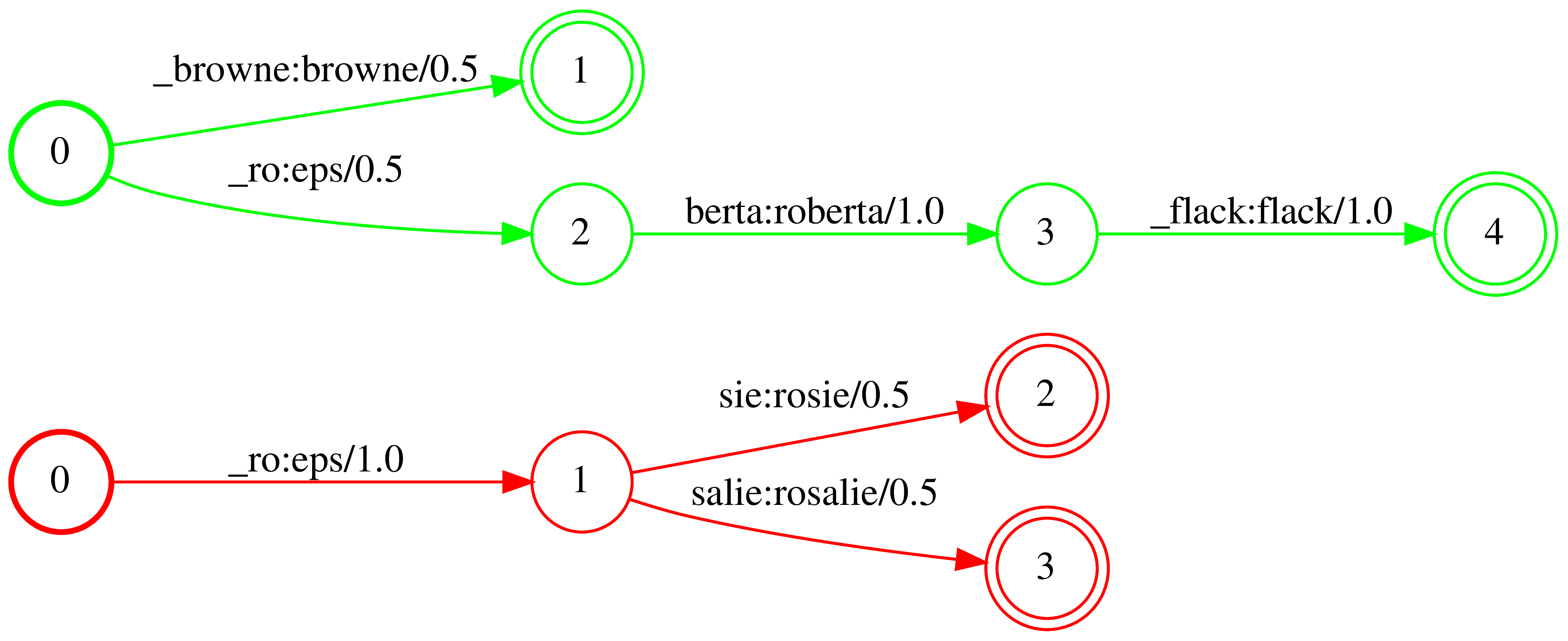}
\vspace{0.2em}
\caption{Component FST for class $@song$ (red) and $@artist$ (green). Note that there is no loop-back from the final states to the start states. The handling of multiple entries into the same class is controlled by the class emission probability, more specifically the neural decider.}
\label{fig:bias_fst}
\end{figure}

Similar to a conventional sequence-to-sequence neural network \cite{pronlearning}, we can represent NFCLM as an infinite FST where new states and arcs are added dynamically as we extend $\mathbf{h^w}$ with new symbols. This implementation allows NFCLM to be used like a normal FST while wrapping the decider, the background neural LM, and multiple non-background component FSTs under the hood. As an illustration, we show how NFCLM consumes the sequence $\{\_play, \_ro, sie, \_by, \_browne\}$ in Fig.~\ref{fig:dfst}. This toy example assumes that we have two classes, $@artist$ and $@song$, with entities \{$(\_ro,berta,\_flack)$, $(\_browne)$\} and \{$(\_ro, sie)$, $(\_ro, salie)$\}, respectively (Fig.~\ref{fig:bias_fst}). At every decoding step, new class alignments are added and pruned away based on the beam setting. 

\section{Experimental Setup}
\label{sec:experimental_setup}

\subsection{Text Datasets and Evaluation Procedure}
\label{sec:asr_test_sets}

On the training side, we have access to a 29M-sentence (147M-word) \emph{background} text corpus sampled from crowdsourced data and the manually transcribed Facebook voice assistant traffic of users who have agreed to having their voice activity reviewed and analyzed. We also have access to a set of CFG rules that represent the music and weather domains. The rules consist of 291 patterns and seven non-terminals, $@album$ (11.8K entities), $@artist$ (10.9K), $@genre$ (1.0K), $@playlist$ (3.8K), $@song$ (89.0K), $@station$ (5.2K), and $@location$ (48.1K). From these data we construct a 27M-sentence (183M-word) \emph{non-background} text corpus by fully expanding the CFG rules with uniformly sampled patterns and entities. Together, these comprise all the available text data for training.


We consider two evaluation sets in this work. The first one, \emph{General} (\texttt{GEN}), consists of 16.0K manually transcribed utterances collected from the voice activity of users who interacted with the Facebook voice assistant and have agreed to having their voice activity reviewed and analyzed. The second evaluation set, \emph{Entity-Heavy} (\texttt{ENT}), contains 10.0K utterances generated with an in-house TTS engine; each utterance contains on average 3.4 regular words and 3.1 entity words. The references are generated randomly by expanding the CFG rules as described above. We use TTS data for evaluation in this work in order to test the proposed method on more diverse patterns and entities compared to those seen in traffic. Note that using TTS data does not change the overall conclusion, only its magnitude. The goal of external LM fusion is to improve WER on \texttt{ENT} while minimizing degradation on \texttt{GEN}, by leveraging all available text data without modifications to the base ASR model.

\subsection{Baseline Systems}

Our base E2E ASR model is a sequence transducer (RNN-T) \cite{Graves12} containing approximately 100M parameters; the overall architecture and training methods are described in detail in \cite{le21_interspeech}. To summarize, the model employs a 28-layer streamable low-latency Emformer encoder \cite{ShiWangWuEtAl21} with a stride of four, 40ms lookahead, and 120ms segment size. The target units are 4095 unigram WordPieces \cite{Kudo2018SubWord} built with SentencePiece \cite{KudoRichardson18}. The model is trained on 1.7M hours of in-house data using AR-RNNT loss \cite{Mahadeokar2021AR-RNNT} and trie-based deep biasing \cite{le21_interspeech}.

We consider two baseline methods to utilize the text data, both based on 1st-pass shallow fusion. The first method (\texttt{NNLM}) encodes all available text data (background, 147M words, and non-background, 183M words, fed in random order) with a single neural LM consisting of three LSTM layers (20.7M parameters), trained using Cross Entropy (CE) loss and 0.1 dropout. Training sentences are tokenized into WordPieces using the same vocabulary as that of the base ASR model. The second method (\texttt{NNLM+FST}) combines a similar neural LM trained \emph{only on the background text} with a generic biasing FST built on the CFG data. This FST is constructed from the CFG patterns and the non-terminals are statically replaced with class-specific FSTs built on the entity lists (e.g., Fig. \ref{fig:bias_fst}), following \cite{le21_interspeech}; the final FST is 416MB in size. With this method, we do not need to re-train the neural LM when new entities are added. Note that unlike \texttt{NFCLM}, \texttt{NNLM+FST} treats neural LM and FST as separate systems rather than as components in a unified framework.




\subsection{NFCLM Setup}

As described in Section \ref{sec:nfclm}, NFCLM is defined by the background LM, class-component FSTs, and the decider. We reuse the background neural LM and class-specific FSTs in \texttt{NNLM+FST} for \texttt{NFCLM} to ensure fair comparison. The decider is a small sequence classification model consisting of a single LSTM layer (286K parameters) trained with CE loss and 0.1 dropout. The training set contains a mixture of background (10\%) and CFG (90\%) data. This strategy exposes the decider to more training data and avoids overfitting to fixed patterns. To re-balance the probabilities produced by the decider, we re-normalize them using the frequency of each output class appearance in the CFG training data; we found that this re-normalization is necessary to prevent the probability mass from being concentrated on the $@bg$ class. Since we do not need to perform static FST replacement for NFCLM, the total size taken up by the class-component FSTs is only 15MB, bringing the total size to 36MB (the neural LMs are 8-bit quantized).

\section{Results and Discussion}
\label{sec:results}
\begin{figure}
\includegraphics[width=\columnwidth]{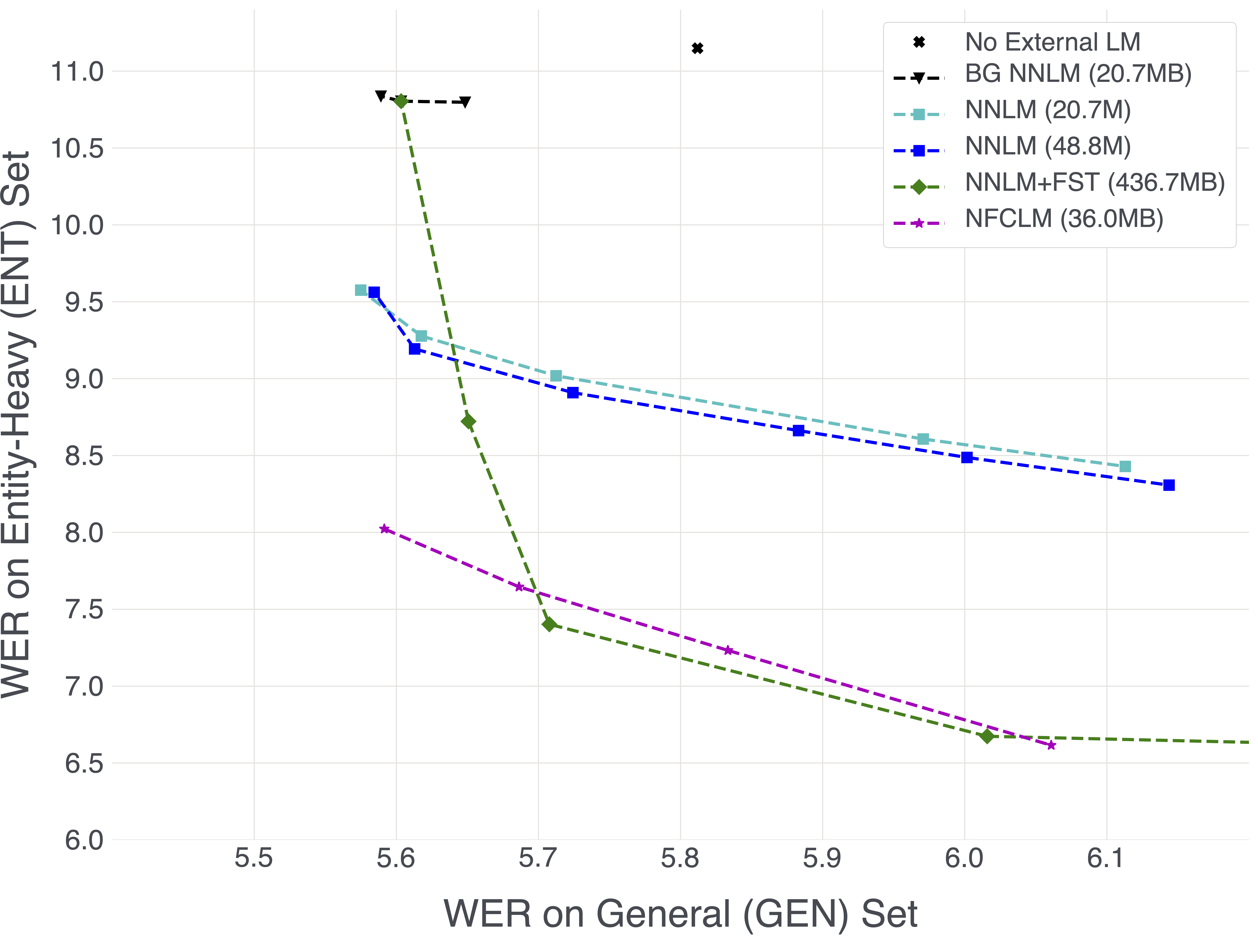}
\caption{WER comparison of different LM integration methods on the \emph{General} (\texttt{GEN}) and \emph{Entity-Heavy} (\texttt{ENT}) evaluation sets.}
\label{fig:wer_nfclm}
\end{figure}

We assess the performance of each LM integration method through its effect on both the \texttt{GEN} and \texttt{ENT} evaluation sets. We first obtain different operating points for each method by sweeping on various hyperparameters, including neural LM weight, internal LM weight, and FST shallow fusion weight (only applicable to \texttt{NNLM+FST}), which control the WER tradeoff between the two sets. We then plot the convex hull corresponding to each collection of operating points in Fig. \ref{fig:wer_nfclm}. We are thus able to visualize and compare the efficient frontiers of different methods; the closer the points are to the origin, the better the method. Note that the vanilla RNN-T result without any external LM is displayed as a single point in the graph, as we cannot control the tradeoff between \texttt{GEN} and \texttt{ENT} in this case.

We can see the benefit of neural LM shallow fusion in Fig. \ref{fig:wer_nfclm} (\texttt{NNLM}), especially on the \texttt{ENT} evaluation set with 15--25\% relative WER reduction (WERR) compared to the vanilla RNN-T baseline. A similar neural LM trained with only the background text (\texttt{BG NNLM}) offers relatively modest improvement on \texttt{ENT}, pointing to the importance of the non-background training text. Increasing the model size from 20.7M to 48.8M parameters does not improve the results significantly. Compared to \texttt{NNLM}, FST can better memorize rare named entities. This is confirmed through \texttt{NNLM+FST} achieving significantly better results on \texttt{ENT}; however, the method incurs greater degradation on \texttt{GEN} due to overbiasing.

The proposed \texttt{NFCLM} clearly outperforms \texttt{NNLM}, achieving 15.8\% WERR on \texttt{ENT} and similar performance on \texttt{GEN}. The improvement here, similar to \texttt{NNLM+FST}, can be attributed to the use of FSTs for class-specific entities. Compared to \texttt{NNLM+FST}, \texttt{NFCLM} is significantly less prone to overfitting. If we want to obtain the best performance on \texttt{ENT} while minimizing the degradation on \texttt{GEN} (leftmost point of each curve), then \texttt{NFCLM} is the clear winner. Looking at the rest of the curve, \texttt{NFCLM} and \texttt{NNLM+FST} achieve very similar operating points; however, \texttt{NFCLM} uses drastically less storage space (36.0MB vs. 436.7MB, 12 times smaller), making it more amenable to be used on-device. The decoding speed, measured in Real-Time Factor (RTF), with \texttt{NFCLM} is slightly worse (0.28 vs. 0.26) due to the extra computation incurred by the multiple class alignments. By factoring out the model into smaller components, we can handle changes in the entity lists and expand to new domains more quickly. When any class-specific entity list changes, we only need to rebuild the FST for that class. Supporting new domains and patterns does require re-training the decider neural network; however, this is considerably cheaper than re-training the entire NNLM since the decider is much more lightweight and does not require as much data to converge. Moreover, by implementing \texttt{NFCLM} as a dynamic FST, no change to the decoder is needed.

\section{Conclusion and Future Work}
\label{sec:conclusions}
In this paper, we proposed NFCLM, a novel method that factorizes the LM into separate components (neural LM and FSTs) in a unified and mathematically consistent framework. We demonstrated that the proposed method, when used in end-to-end ASR, significantly outperforms vanilla NNLM in modeling rare named entities, and is less prone to overbiasing and much more compact than the traditional FST shallow fusion approach.

For future work, we plan on carrying out more in-depth studies to better understand how robust NFCLM is to more varied use cases in terms of entity types, entity test set sizes, and train/test data sets. We also plan to expand the data on which the decider is trained. Instead of focusing on a limited set of manually annotated patterns, we plan to use a large corpus of text and use an entity tagger to construct patterns and entity lists automatically. We will also explore alternative LM architectures beyond LSTMs.

\newpage
\footnotesize
\bibliographystyle{IEEEbib}
\bibliography{refs}

\end{document}